%% file: main.tex
\documentclass[10pt,twocolumn,letterpaper]{article}

\usepackage[pagenumbers]{cvpr} % To force page numbers, e.g. for an arXiv version

\usepackage{multirow}
\usepackage{colortbl}

\input{preamble}
\definecolor{cvprblue}{rgb}{0.21,0.49,0.74}
\usepackage[pagebackref,breaklinks,colorlinks,allcolors=cvprblue]{hyperref}

\def\confName{CVPR}
\def\confYear{2027}

\title{RemoteEvolver: Expanding Earth Observation Capabilities\\ through Agent Harness Evolution}

\author{%
  \textbf{Shengxiang Xu}\textsuperscript{1,3\,*},
  \textbf{Liang Yao}\textsuperscript{2\,*},
  \textbf{Shuyi Chen}\textsuperscript{2},
  \textbf{Yishu Wang}\textsuperscript{1},
  \textbf{Shimin Di}\textsuperscript{1\,\textdagger},
  \textbf{Fan Liu}\textsuperscript{2\,\textdagger},\\
  \textbf{Bishun Yao}\textsuperscript{2},
  \textbf{Hanghui Guo}\textsuperscript{1},
  \textbf{Xinyu Liu}\textsuperscript{3},
  \textbf{Chengye Liang}\textsuperscript{1},
  \textbf{Yuyu Luo}\textsuperscript{3},
  \textbf{Min-Ling Zhang}\textsuperscript{1}\\[0.4em]
  {\textsuperscript{1} Southeast University\quad
   \textsuperscript{2} Hohai University}\\
  {\textsuperscript{3} The Hong Kong University of Science and Technology (Guangzhou)}
}

\begin{document}
\maketitle

\begingroup
\renewcommand{\thefootnote}{\fnsymbol{footnote}}
\footnotetext[1]{Equal contribution. This work was done while Shengxiang Xu was a visiting research student at HKUST(GZ).}
\footnotetext[2]{Corresponding authors: Shimin Di (shimin.di@seu.edu.cn), Fan Liu (fanliu@hhu.edu.cn)}
\endgroup

\input{0_Abstract}    
\input{1_Introduction}
\input{2_RelatedWork}

\input{3_Method}

\input{4_Experiment}
\input{5_limitation}
\input{6_Conclusion}
\input{7_LLM}
{
    \small
    \bibliographystyle{ieeenat_fullname}
    \bibliography{main}
}

% WARNING: do not forget to delete the supplementary pages from your submission 
% \input{sec/X_suppl}

\end{document}

%% file: 0_Abstract.tex
\begin{abstract}
We are interested in developing a continually self-improving remote sensing agentic system. It should identify and address its capability gaps, retaining the resulting improvements to enhance performance on subsequent tasks.
% 第三句太表象了，没有深层的探究他们为什么不行。应该是：现有的方法由于他们XXXX，无法满足这个需求。
However, existing systems typically rely on fixed toolsets and confine their experience accumulation to re-orchestrating predefined capabilities, constraining both capability expansion and the transfer of improvements across tasks.
% 然后说为什么为什么要做Harness自进化。例如：harness作为agentic systems的基础设施，它有利于xxx（承接我们第二句话）
% 为什么作为基础就可以自进化，少了一层逻辑，比如他很好更新不像模型那样需要不断训练？
As the infrastructure for agentic systems, the harness manages tools and workflow knowledge as editable, persistent resources outside the base model. These resources can therefore be updated and reused across tasks without additional post-training.
% 然后基于上述观察，我们xxx后面就不用改了
Based on these observations, we propose RemoteEvolver, an external harness for remote sensing agents that drives tool and workflow evolution through execution feedback. With the base model parameters fixed, RemoteEvolver acquires missing specialized tools on demand and distills verified successful execution experience from individual tasks into agent skills that can be reused across tasks. The acquired tools and accumulated skills jointly support planning and execution on subsequent tasks. Experiments on Earth-Bench and OpenEarthAgent-Bench show that RemoteEvolver achieves average accuracy gains of 8.18 and 23.86 percentage points over its initial state when tools are fully available and partially missing. Under the same settings, it also outperforms the state-of-the-art methods by an average of 6.77 and 17.10 percentage points, respectively.
The code is publicly available at \url{https://github.com/XuShengXianggg/RemoteEvolver}.
\end{abstract}

%% file: 1_Introduction.tex
\section{Introduction}
\begin{figure}[t]
    \centering
    \includegraphics[width=\columnwidth]{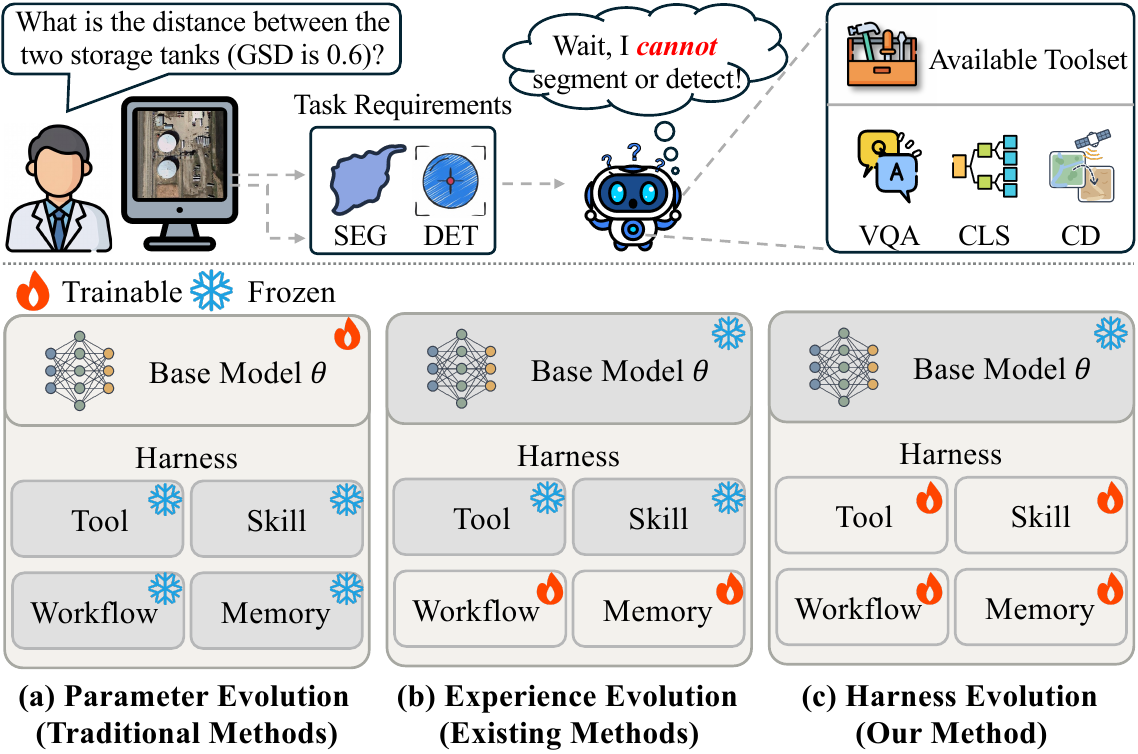}
    \caption{
        Motivation of RemoteEvolver. \textbf{Top}: Remote sensing agents encounter unbridgeable capability gaps when required specialized tools are missing from their predefined toolsets. \textbf{Bottom}: Comparison of existing remote sensing agent evolution paradigms.
    }
    \label{fig:motivation}
\end{figure}

% 遥感智能体为什么需要自进化？
% 介绍遥感智能体可以做什么
Remote sensing agents autonomously plan and invoke perception models and geospatial analysis tools to perform Earth observation tasks~\cite{xu2026rs,hu2026ringmo,weiss2025mapbot,talemi2026agentic,tang2026intelligent} according to users’ queries.
% 实际部署面临什么问题，现实需求和系统能力存在什么gap
However, user requests sometimes exceed a system’s current capabilities in practical deployment. 
As illustrated in Fig.~\ref{fig:motivation}, an agent may understand a request~\cite{yao2026remoteagent,kao2026towards,xu2025robustflow,mu2026hirs} and formulate a reasonable plan~\cite{chen2025cangling,yao2026remotereasoner,xiao2026geoforge,ren2026flowevo}, yet still fail to complete the task because a required specialized tool is unavailable.
% 所以需要让遥感智能体学会自我进化
To support users’ analytical queries over time, remote sensing agents must develop self-evolving capabilities, autonomously identifying and addressing capability gaps during execution, and retaining verified improvements for reuse in subsequent tasks.

% 现有方法为什么还不够？
% 现有方法已经做出了哪些改进
To achieve self-evolution, recent remote sensing agentic systems have explored various strategies, ranging from workflow repair and experience accumulation~\cite{wu2026rsmem,chen2025cangling,xiao2026geoforge,dai2026experience} to on-demand script generation by proactively consulting documentation~\cite{zhao2026openearth}. 
% 但这些改进在什么情况下仍然不足
Despite this progress, these approaches primarily focus on optimizing their internal logic or generating simple rule-based scripts. 
Such script-level adaptations remain fundamentally constrained when task requirements exceed the perception capabilities supported by the agent's existing toolset. 
Workflow adjustment and code generation alone cannot provide specialized capabilities that require pretrained models. 
To bridge these gaps, agents need to autonomously discover, acquire, and integrate pretrained models from external resources.

% 为什么我们要通过harness实现自进化？

% ------------
% 这里和摘要对齐，先说harness的易变性，所以说他天生比RemoteZero这种需要更新模型参数的方法要好。不要这么刻意的上来就说zero，你甚至可以不提他的名字
% ------------

% 承接上文，引入harness
To effectively integrate these newly acquired models and corresponding practical experiences, agents can leverage their \textit{harness}~\cite{guo2026drevo}, which is the runtime support layer connecting the base model to the execution environment.
% harness有哪些特质，为自演化带来了哪些优势
As the infrastructure for agentic systems in Fig.~\ref{fig:motivation}, the harness manages tools and workflow knowledge as editable, persistent resources outside the base model.
Consequently, modifying harness allows the system to seamlessly archive and reuse acquired capabilities across tasks.
% 自然而然的和其他通过参数更新的演化方法形成对比
This external integration inherently bypasses the continuous training overhead required by approaches that attempt to internalize new capabilities through model parameter optimization~\cite{yao2026remotezero, xin2026segearth}.

Motivated by these observations, we propose RemoteEvolver, a self-evolving harness for remote sensing agents that connects capability acquisition with the accumulation of reusable experience. With the base model parameters frozen, it supports continued capability growth through two complementary mechanisms: dynamic toolset expansion and skill library consolidation. Dynamic toolset expansion enables agents to autonomously acquire and integrate pretrained models from open-source repositories as task requirements evolve. Skill library consolidation distills verified successful executions into reusable procedures that guide subsequent tasks. Crucially, successful workflows involving newly acquired tools can also be consolidated into reusable skills, allowing new capabilities to be retained together with the procedural knowledge needed to apply them. It enables remote sensing agents to build on both advances in the external model ecosystem and their own accumulated execution experience across tasks.

We conduct a comprehensive evaluation of RemoteEvolver on Earth-Bench~\cite{feng2026earth} and OpenEarthAgent-Bench~\cite{shabbir2026openearthagent} with fixed base model parameters. By actively evolving its external harness, the system achieves average accuracy improvements of 23.86$\%$ and 8.18$\%$ over its initial state when tools are partially missing and fully available, respectively. Besides, under the same settings, RemoteEvolver significantly outperforms state-of-the-art methods by an average of 17.10$\%$ and 6.77$\%$. Furthermore, extensive analysis confirms that this evolution paradigm generalizes well across various mainstream foundation models, consistently delivering strong transferability on diverse remote sensing tasks.

In summary, our contributions are as follows:

\begin{itemize}
    \item We introduce a harness-centric self-evolution paradigm for Earth observation agents, enabling continual capability growth without updating the base model parameters.

    \item We propose RemoteEvolver, a practical framework that enables the agentic system to incorporate emerging specialized models into agent systems and build upon accumulated task experience across tasks.

    \item Extensive evaluations demonstrate that RemoteEvolver significantly outperforms state-of-the-art methods under both complete and constrained tool settings.
\end{itemize}

%% file: 2_RelatedWork.tex
\section{Related Work}
\begin{figure*}[t]
    \centering
    \includegraphics[width=\textwidth]{
    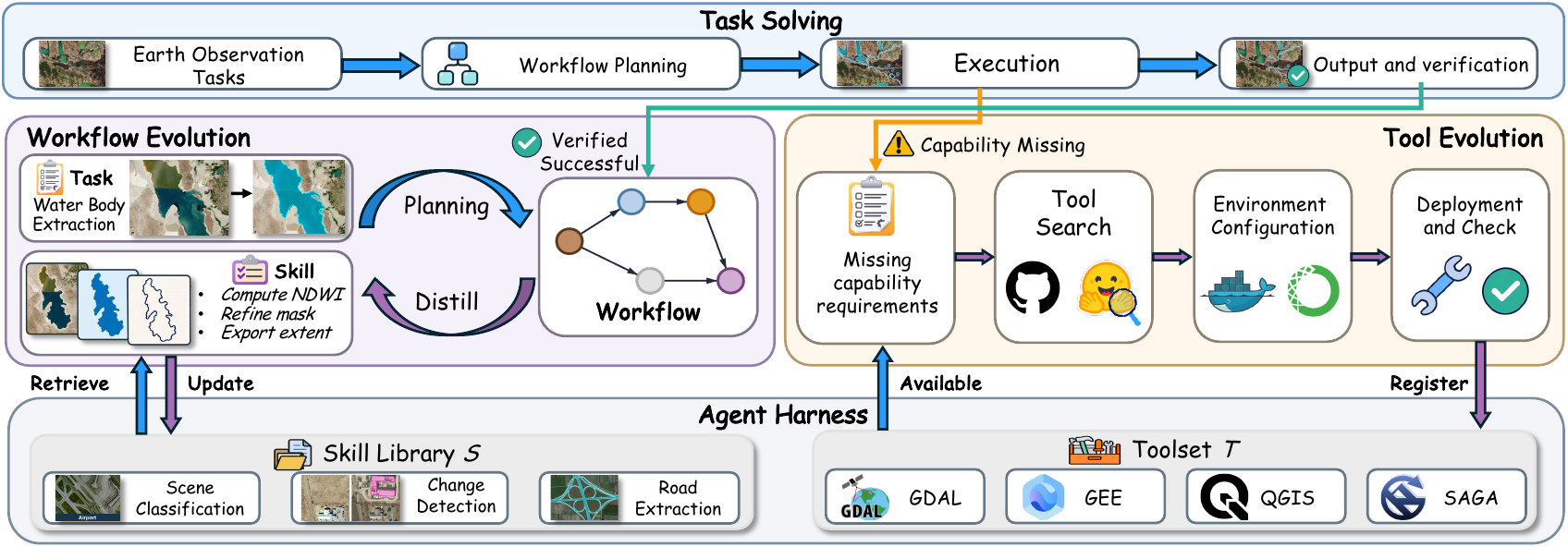
    }
    \caption{
        Overview of RemoteEvolver. When encountering capability gaps during task solving, the tool evolution module autonomously searches, configures, and deploys missing external resources into the available toolset (Dynamic Toolset Expansion). Concurrently, the workflow evolution module distills verified successful execution workflows into reusable procedural skills (Skill Library Consolidation).
    }
    \label{fig:method}
\end{figure*}

\subsection{Model-Centric Earth Observation Systems}
Advances in pretraining have broadened the capabilities of model-centric Earth observation systems in representation learning, semantic understanding, and spatial prediction. For instance, SatMAE~\cite{cong2022satmae} learns transferable representations from temporal or multispectral satellite imagery. Broader input coverage is achieved through optical–SAR imagery fusion in CROMA~\cite{fuller2023croma} and SkySense~\cite{guo2024skysense}, with AnySat~\cite{astruc2025anysat} further unifying data across resolutions, scales and modalities. Language-based capabilities have expanded from image–text alignment, as illustrated by RemoteCLIP~\cite{liu2024remoteclip}, to richer forms of interaction. These include image-level and region-level dialogue in GeoChat~\cite{kuckreja2024geochat} and interactive understanding of multi-sensor and multi-temporal observations in EarthDial~\cite{soni2025earthdial}. Spatial prediction has also seen parallel advances toward more flexible outputs. RemoteSAM~\cite{yao2025remotesam} unifies multiple perception tasks around referring expression segmentation, whereas GeoGround~\cite{zhou2024geoground} represents horizontal and oriented bounding boxes and segmentation masks in a common format. GeoPixel~\cite{shabbir2025geopixel} further connects conversational understanding with pixel-level segmentation. By structuring these versatile capabilities into an autonomous reasoning workflow, RemoteReasoner~\cite{yao2026remotereasoner} bridges the gap between foundational perception and end-to-end task execution. Together, these developments provide a growing range of remote sensing capabilities. With appropriate task adaptation, the resulting models can be exposed through callable interfaces as specialized tools for agents tackling complex tasks.

\subsection{Agentic Earth Observation Systems}
Agentic Earth observation systems coordinate domain-specific models and processing tools to translate user requests into executable workflows. RS-Agent~\cite{xu2026rs} and Earth-Agent~\cite{feng2026earth} combine specialized tools with domain knowledge to solve remote sensing tasks, while RemoteAgent~\cite{yao2026remoteagent} routes requests between internal models and external experts according to task requirements. To support workflow construction and adaptation, OpenEarth-Agent~\cite{zhao2026openearth} generates task-specific tool scripts and iteratively repairs workflows. Earth-Agent-Pro~\cite{lv2026earth} uses domain skills and execution evidence to repair affected downstream steps, while RSMeM~\cite{wu2026rsmem} reflects on failures to improve subsequent attempts on the same task. Further work explores experience-driven approaches to improving planning and execution reliability. CangLing-KnowFlow~\cite{chen2025cangling} consolidates successfully adapted workflows into reusable templates and derives repair rules from failures. GeoEvolver~\cite{dai2026experience} accumulates tool-use experience from successful and failed exploration trajectories. GeoForge~\cite{xiao2026geoforge} organizes execution knowledge at multiple levels through workflow graphs, action-level experience, and standard operating procedures for skills. Despite these advances, a critical limitation persists: these systems primarily rely on pre-integrated tools, limiting their ability to autonomously handle tasks that require integrating additional specialized external tools.

%% file: 3_Method.tex
\section{RemoteEvolver}
We propose RemoteEvolver in Fig.~\ref{fig:method}, which integrates both tool evolution and workflow evolution as two complementary components. Specifically, these two core components are realized through dynamic toolset expansion and skill library consolidation, respectively. By continually updating the external harness while keeping the base model parameters fixed, it enables agents to accumulate and reuse new capabilities across diverse remote sensing tasks.

\subsection{Problem Formulation}
We consider a sequence of remote sensing tasks $\{z_t=(q_t,x_t)\}_{t=1}^{N}$, where $q_t$ denotes a natural-language instruction and $x_t$ denotes the associated input data. Let $M_\theta$ be the base model, whose parameters $\theta$ remain fixed throughout evolution. To achieve capability accumulation independent of model parameters, we represent the evolving persistent state of the external harness as
\begin{equation}
    H_t=(\mathcal T_t,\mathcal S_t),
\end{equation}
where $\mathcal T_t$ is the available toolset providing executable operations, and $\mathcal S_t$ is a library of skill documents acting as the system's long-term memory to provide procedural knowledge. Together, they support the dynamic generation of task-specific reasoning workflows.

Given task $z_t$, the agent plans an initial workflow $W_t^{(0)}$ and execute step-by-step based on the current harness $H_t$. Crucially, if a capability bottleneck or unsatisfactory intermediate outcome is encountered during execution, the agent can on-demand acquire missing tools, updating the harness to an intermediate state $\widetilde H_t$ and dynamically adapting the workflow to resume the task. If no tool acquisition is triggered, the harness remains unchanged ($\widetilde H_t = H_t$). This interleaved planning and execution process is formulated as:
\begin{equation}
\begin{aligned}
    W_t^{(0)} &= \operatorname{Plan}_{\theta}(z_t; H_t), \\
    (\hat y_t, \tau_t, W_t, \widetilde H_t) &=\operatorname{Execute}_{\theta}(z_t,W_t^{(0)}; H_t),
\end{aligned}
\end{equation}
where $W_t^{(0)}$ is the initial workflow generated by the base model, $W_t$ is the final executed workflow, $\hat y_t$ is the task output, and $\tau_t$ is the detailed execution trace. Given verification feedback $f_t$ provided by a external evaluator, the harness is subsequently updated as
\begin{equation}
    H_{t+1} =\operatorname{Update}(\widetilde H_t, W_t, z_t, \tau_t, f_t).
\end{equation}
This evolutionary process is driven by a complementary mechanism: registering verified new tools into the toolset and distilling successful workflow $W_t$ into reusable skills.

Our objective is to push capability boundaries and improve performance on subsequent tasks through continuous updates to the external harness, without modifying parameters $\theta$. During evaluation, the harness state $H_t$ is strictly frozen, and test feedback is excluded from evolution.

\subsection{Dynamic Toolset Expansion}
When the current toolset $\mathcal T_t$ lacks a required remote sensing operation or fails to yield satisfactory execution outcomes, RemoteEvolver searches GitHub and HuggingFace for relevant implementations and pretrained models. This search process is driven by the capability gap between the current task $z_t$ and the available toolset $\mathcal T_t$. To select resources for deployment, the system matches the functional requirements and IO constraints against candidate resources, such as \texttt{README} files or model documentation.
This selection process specifically evaluates supported operations, applicable data modalities, and output formats. 
Let the initially retrieved candidate tool set be:
\begin{equation}
    \mathcal U_t = \operatorname{Search}(z_t, \mathcal T_t).
\end{equation}

For each selected resource, the system retrieves code, prepares the runtime environment, and constructs a callable interface. For tools based on pretrained models, it also automatically downloads and deploys the required weights. We represent each deployed candidate tool $u \in \mathcal U_t$ as:
\begin{equation}
    u=(r,\omega,\mathcal E,g),
\end{equation}
where $r$ denotes the retrieved implementation, $\omega$ represents the pretrained weights, $\mathcal E$ signifies the runtime environment, and $g$ stands for the interface. The interface is responsible for receiving task data and operational parameters, invoking the underlying implementation, and returning results. For tools that do not rely on pretrained models, we set $\omega=\varnothing$.

Each candidate tool $u \in \mathcal U_t$ undergoes an execution check. If an invocation fails, the system uses the returned error information to revise the interface adapter code and repeats the check. Let $\mathcal V_{\mathrm{exec}}(u)\in\{0,1\}$ denote the final execution check outcome. The set of accepted tools is:
\begin{equation}
    \mathcal U_t^{\mathrm{pass}}=\left\{u\in\mathcal U_t\mid\mathcal V_{\mathrm{exec}}(u)=1\right\}.
\end{equation}
Tools that pass the execution check are registered in the toolset, yielding the updated available toolset:
\begin{equation}
    \widetilde{\mathcal T}_t=\mathcal T_t\cup\mathcal U_t^{\mathrm{pass}}.
\end{equation}
The newly registered specialized tools are immediately available to support the current task's execution, enabling the dynamic adaptation of the workflow, and persistently retained for reuse in subsequent tasks.

\subsection{Skill Library Consolidation}
Tool acquisition expands the underlying operations available to the system. To reuse execution experience, RemoteEvolver distills verified workflows into \texttt{skill.md} documents and preserves them in a persistent skill library. These documents explicitly guide and condition subsequent workflow planning by providing detailed structured step constraints and tool dependencies.

Given a task $z_t$, the system retrieves a set of relevant skills $\mathcal K_t$ from the current skill library $\mathcal S_t$:
\begin{equation}
    \mathcal K_t = \operatorname{Retrieve}_{\theta}(z_t, \mathcal S_t).
\end{equation}
The planner would read these documents and combines their structured guidance with the current remote sensing data and task requirements to construct and dynamically adapt the initial execution workflow:
\begin{equation}
    W_t^{(0)}=\operatorname{Plan}_{\theta}(z_t;\mathcal T_t,\mathcal K_t),
\end{equation}
where $\mathcal T_t$ denotes the toolset currently available during the initial planning phase. The retrieved skills guide the selection of processing steps and parameters. If no applicable skill is found, we set $\mathcal K_t=\varnothing$, and the planner directly generates a workflow from the task requirements and the existing toolset. If tool evolution is triggered during execution, the system leverages the newly acquired tools to dynamically adapt and finalize the workflow into $W_t$.

After task completion, the system performs a closed-loop update of the skill library using the final workflow $W_t$, execution trace $\tau_t$, and task feedback $f_t$. Verified successful workflows are distilled into new skills by the base model $\theta$. Concurrently, existing skills are revised, merged, or deactivated based on their historical reuse outcomes, which are tracked as success rates and invocation counts. Deactivated skills will no longer participate in subsequent planning stages. This update process can be expressed as:
\begin{equation}
    \mathcal S_{t+1}=\operatorname{Update}_{\theta}(\mathcal S_t, W_t, z_t, \tau_t, f_t).
\end{equation}
If newly acquired tools are invoked in a successful workflow, their complete usage procedures are also retained within the generated skill documents, thereby directly supporting future workflow planning through retrieval.

%% file: 4_Experiment.tex
\section{Experiments}

\subsection{Experimental Setup}
\noindent
\textbf{Datasets.} We evaluate RemoteEvolver on Earth-Bench~\cite{feng2026earth} and OpenEarthAgent-Bench~\cite{shabbir2026openearthagent}. We split tasks into evolution and test sets at a ratio of 7:3. When splitting, we aim to keep task category distribution similar in both sets.

\noindent
\textbf{Baselines.} We compare RemoteEvolver with 5 representative remote sensing agent systems, including RSMeM~\cite{wu2026rsmem}, RS-Agent~\cite{xu2026rs}, Earth-Agent~\cite{feng2026earth}, OpenEarthAgent~\cite{shabbir2026openearthagent}, and RemoteAgent~\cite{yao2026remoteagent}. We additionally include our initial system RemoteEvolver$^*$ as an internal baseline.

\noindent
\textbf{Implementation Details.} In the main experiments, we use Qwen3.8 27B~\cite{qwen38} as the backbone for all methods except OpenEarthAgent and RemoteAgent, which retain their officially released checkpoints. We further evaluate RemoteEvolver with GPT-5.6 Luna, Qwen3.8 Max~\cite{qwen38}, DeepSeek V4.1 Flash~\cite{xu2026deepseek}, MiniMax M3~\cite{lai2026minimax}, and GLM 5.3~\cite{zeng2026glm}. Within each method, the same model serves as both the planner and executor. All models operate in thinking mode with the temperature set to 0. We impose no explicit limits on execution steps or output tokens. The tool execution environment is equipped with 8 NVIDIA RTX 4090 GPUs. More implementation details are provided in Appendix.

\subsection{Evaluation Protocol}
\noindent
\textbf{Tool Availability.}
We consider two tool settings. \textit{Tool-Available} initially provides a complete toolset covering all tasks, whereas \textit{Tool-Missing} initially provides a predefined subset of the full toolset, consisting of basic remote sensing tools. Detailed tool lists are provided in Appendix.

\noindent
\textbf{Evaluation Procedure.}
We divide evolution process into 8 stages and save the harness state $H_t$ after each stage, where $t=0,1,\dots,8$. We evaluate all harness states on the same testset. The harness state remains frozen during evaluation, and test feedback is not used for subsequent evolution.

\noindent
\textbf{Evaluation Metrics.}
We evaluate system evolutionary process in terms of task performance, transferability and retention capability. We measure task performance with accuracy $\mathrm{Acc}(H,D)=\frac{1}{|D|}\sum_{q\in D}a_q$, where $H$ denotes the harness state, $D$ denotes the testset, and $a_q \in \{0, 1\}$ represents the binary correctness of task $q \in D$.

To measure the system's transferability, we further split the testset $D$ into the near subset $D_\mathrm{Near}$ and the far subset $D_\mathrm{Far}$. The near subset $D_\mathrm{Near}$ contains tasks similar to those in evolution, while the far subset $D_\mathrm{Far}$ contains tasks unseen during evolutionary process. Then, we utilize forward transfer (FWT) to measure the transferability gain of the current state $H_t$ over the initial state $H_0$:
\begin{equation}
    \mathrm{FWT_{g}^{(t)}} = \mathrm{Acc}(H_t,D_g)-\mathrm{Acc}(H_0,D_g),
\end{equation}
where $g\in\{\mathrm{Near},\mathrm{Far}\}$ and $D_g$ denotes the corresponding test subset. Higher FWT indicates better transferability.

We also measure retention capability as the proportion of previously solved tasks that current system fails to solve. For each task $q\in D$, let $a_q^{(t-1)}$ and $a_q^{(t)}$ denote its binary correctness on the last attempt and under the current harness state $H_t$, respectively. We define Retention Rate (RR) as:
\begin{equation}
    \mathrm{RR}_t = \frac{\sum_{q\in D} \mathbb{I}(a_q^{(t-1)} = 1 \land a_q^{(t)} = 0)}{\sum_{q\in D} \mathbb{I}(a_q^{(t-1)} = 1)},
\end{equation}
where $\mathbb{I}(\cdot)$ denotes the indicator function. A lower RR indicates better retention capability.

\begin{table}[t]
    \centering
    \caption{
        Comparison of task accuracy ($\%, \uparrow$) on Earth-Bench~\cite{feng2026earth} (EB) and OpenEarthAgent-Bench~\cite{shabbir2026openearthagent} (OEAB) under available and missing tools. * denotes the initial system before evolution. \textbf{Best} and \underline{second-best} results are highlighted. Values in parentheses indicate the accuracy improvements achieved by evolution. 
    }
    \label{tab:performance_comparison}

    \begingroup
    \footnotesize
    \setlength{\tabcolsep}{3pt}
    \renewcommand{\arraystretch}{1.15}

    % 数值与增益组成一个水平、垂直均居中的单元格
    \newcommand{\scoregain}[2]{%
        \begin{tabular}[c]{@{}c@{}}
            \textbf{#1} \\[-1pt]
            {\scriptsize $(+#2)$}
        \end{tabular}%
    }

    \resizebox{\columnwidth}{!}{%
        \begin{tabular}{lccccc}
            \toprule

            \multirow{2}{*}{\textbf{Methods}}
            & \multirow{2}{*}{\textbf{Publication}}
            & \multicolumn{2}{c}{\textbf{Tool-Available}}
            & \multicolumn{2}{c}{\textbf{Tool-Missing}} \\

            \cmidrule(lr){3-4}
            \cmidrule(lr){5-6}

            & & EB & OEAB & EB & OEAB \\
            \midrule

            RSMeM~\cite{wu2026rsmem}
            & ACL'26
            & 45.71
            & \underline{31.91}
            & \underline{25.71}
            & 7.45 \\

            RS-Agent~\cite{xu2026rs}
            & SCIS'26
            & 41.43
            & 25.53
            & 17.14
            & 12.77 \\

            Earth-Agent~\cite{feng2026earth}
            & ICLR'26
            & 48.57
            & 28.72
            & 21.43
            & 5.32 \\

            OpenEarthAgent~\cite{shabbir2026openearthagent}
            & ECCV'26
            & 12.86
            & 10.70
            & 2.86
            & 2.13 \\

            RemoteAgent~\cite{yao2026remoteagent}
            & TGRS'26
            & 4.29
            & 10.64
            & 4.29
            & \underline{18.09} \\

            \midrule

            RemoteEvolver$^*$
            & --
            & \underline{50.00}
            & 27.66
            & 18.57
            & 11.70 \\

            \rowcolor{blue!6}
            RemoteEvolver
            & --
            & \scoregain{55.71}{5.71}
            & \scoregain{38.30}{10.64}
            & \scoregain{47.14}{28.57}
            & \scoregain{30.85}{19.15} \\

            \bottomrule
        \end{tabular}%
    }

    \endgroup
\end{table}

\subsection{Main Results}
Initial tool limitations expose a critical vulnerability in static remote sensing agentic systems. As shown in Table~\ref{tab:performance_comparison}, transitioning from the \textit{Tool-Available} to the \textit{Tool-Missing} setting causes a catastrophic performance drop across all evaluated baselines, with an average accuracy degradation of 14.32 percentage points. For instance, Earth-Agent's accuracy on Earth-Bench drops from 48.57$\%$ to 21.43$\%$, highlighting its rigid dependence on pre-integrated toolsets.

RemoteEvolver fundamentally overcomes this bottleneck through dynamic harness evolution. Under the constrained \textit{Tool-Missing} setting, our system autonomously expands its toolset and distills execution experience, achieving an average accuracy gain of 23.86 percentage points over its initial state. Remarkably, RemoteEvolver operating with initially missing tools (e.g., achieving 47.14$\%$ on Earth-Bench) can approach or even surpass the performance of the strongest external baselines equipped with the complete toolsets. This demonstrates the system's strong resilience when facing severe capability gaps.

Furthermore, the benefits of harness evolution extend beyond merely bridging gaps. It also continuously pushes the performance upper bound. Even under the \textit{Tool-Available} setting where no new tools are required, RemoteEvolver continues to optimize its procedural workflows, yielding an additional average gain of 8.18 percentage points. On Earth-Bench, while our initial system already establishes a strong baseline at 50.00$\%$, the evolutionary process further elevates the state-of-the-art accuracy to 55.71$\%$.

\begin{table}[h!]
    \centering
    \caption{
        Ablation study of tool and skill evolution mechanisms under different tool conditions. Tool evolution is naturally disabled under the \textit{Tool-Available} setting. $\Delta \mathrm{Acc}$ denotes the accuracy gain over the initial system without any evolution. 
    }
    \label{tab:evolution_ablation}

    \begingroup
    \footnotesize
    \setlength{\tabcolsep}{3pt}
    \renewcommand{\arraystretch}{1.2}

    % 跨两行、水平和垂直均居中的指标表头
    \newcommand{\metrichead}[2]{%
        \multirow{2}{*}{%
            \begin{tabular}[c]{@{}c@{}}
                #1 \\[-1pt]
                #2
            \end{tabular}%
        }%
    }

    \resizebox{\columnwidth}{!}{%
        \begin{tabular}{cccccccc}
            \toprule

            \multirow{2}{*}{\textbf{Benchmark}}
            & \multicolumn{2}{c}{\textbf{Evolution}}
            & \metrichead{Acc}
                         {$(\%, \uparrow)$}
            & \metrichead{$\Delta$Acc}
                         {$(\%, \uparrow)$}
            & \metrichead{FWT\textsubscript{Near}}
                         {$(\%, \uparrow)$}
            & \metrichead{FWT\textsubscript{Far}}
                         {$(\%, \uparrow)$}
            & \metrichead{RR}
                         {$(\%, \downarrow)$} \\

            \cmidrule(lr){2-3}
            & Tool & Skill & & & & & \\

            \midrule

            % 仅分区标题行添加灰底
            \rowcolor{blue!6}
            \multicolumn{8}{c}{\textbf{(a) Tool-Missing}} \\

            \midrule

            \multirow{4}{*}{Earth-Bench}
            & $\times$ & $\times$
            & 18.57 & -- & -- & -- & -- \\

            & $\times$ & $\checkmark$
            & 24.29 & $+5.72$ & 3.57 & 7.14 & 0.00 \\

            & $\checkmark$ & $\times$
            & 45.71 & $+27.14$ & 33.93 & 7.14 & 16.67 \\

            & $\checkmark$ & $\checkmark$
            & 47.14 & $+28.57$ & 32.14 & 14.29 & 23.08 \\

            \midrule

            \multirow{4}{*}{%
                \begin{tabular}[c]{@{}c@{}}
                    OpenEarth \\
                    Agent-Bench
                \end{tabular}%
            }
            & $\times$ & $\times$
            & 11.70 & -- & -- & -- & -- \\

            & $\times$ & $\checkmark$
            & 21.28 & $+9.58$ & $4.82$ & $-9.09$ & 46.67 \\

            & $\checkmark$ & $\times$
            & 26.60 & $+14.90$ & 10.84 & 9.09 & 45.95 \\

            & $\checkmark$ & $\checkmark$
            & 30.85 & $+19.15$ & 14.46 & 0.00 & 45.83 \\

            \midrule

            \rowcolor{blue!6}
            \multicolumn{8}{c}{\textbf{(b) Tool-Available}} \\

            \midrule

            \multirow{2}{*}{Earth-Bench}
            & -- & $\times$
            & 50.00 & -- & -- & -- & -- \\

            & -- & $\checkmark$
            & 55.71 & $+5.71$ & 7.14 & 0.00 & 13.33 \\

            \midrule

            \multirow{2}{*}{%
                \begin{tabular}[c]{@{}c@{}}
                    OpenEarth \\
                    Agent-Bench
                \end{tabular}%
            }
            & -- & $\times$
            & 27.66 & -- & -- & -- & -- \\

            & -- & $\checkmark$
            & 38.30 & $+10.64$ & 10.84 & 9.09 & 23.08 \\

            \bottomrule
        \end{tabular}%
    }

    \endgroup
\end{table}

\subsection{Ablation Study}
We conduct comprehensive ablation studies to disentangle the individual contributions of tool and skill evolution. As shown in Table~\ref{tab:evolution_ablation}, while both mechanisms independently improve system performance, their effectiveness and degree of contribution depend heavily on initial tool constraints.

Under the \textit{Tool-Missing} setting, bridging underlying capability gaps serves as the primary prerequisite for evolution. Enabling only skill evolution yields limited gains, improving accuracy by merely 5.72 and 9.58 percentage points on Earth-Bench and OpenEarthAgent-Bench, respectively. It indicates that accumulated procedural experience alone cannot compensate for the absence of core perception models. Conversely, enabling only tool evolution delivers substantial performance leaps, boosting accuracy by 27.14 and 14.90 percentage points across the two benchmarks.

Once tool constraints are lifted, the independent value of skill evolution becomes evident. Under the \textit{Tool-Available} setting, where tool evolution is naturally disabled, continuously distilling skills still yields accuracy gains of 5.71 and 10.64 percentage points across 2 benchmarks. It demonstrates that equipping an agent with complete toolsets does not equate to optimal problem-solving capability. Instead, the system must continuously refine its tool orchestration and invocation logic through accumulated experience.

% Ultimately, integrating both mechanisms enables RemoteEvolver to achieve the best overall performance, validating the strong synergy between expanding capability boundaries and optimizing execution paths.
% Ultimately, integrating both mechanisms into the full RemoteEvolver achieves the best overall performance, reaching 47.14$\%$ and 30.85$\%$ accuracy when tools are partially missing. These results validate the strong synergistic complementarity between tool evolution which expands underlying capability boundaries, and skill evolution which optimizes multi-step execution paths across tasks.

\begin{figure}[t]
    \centering
    \includegraphics[width=\columnwidth]{
        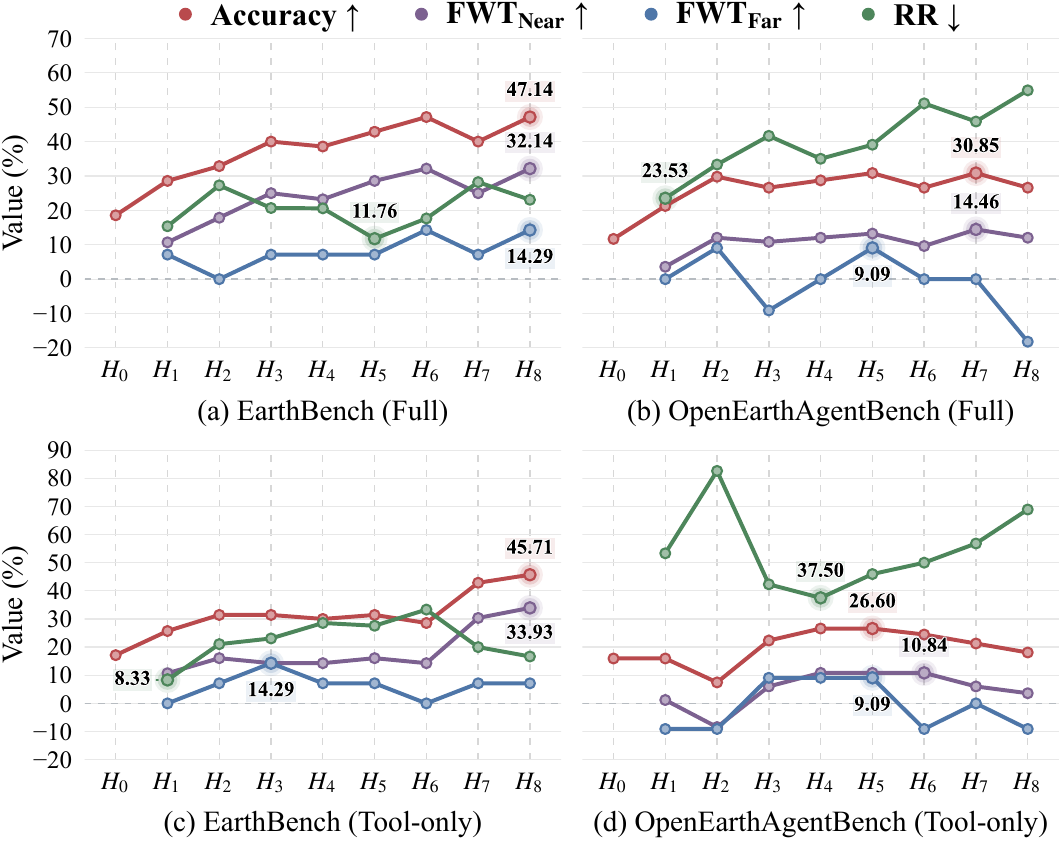
    }
    \caption{
        Evolution trends of RemoteEvolver with full evolution (\textbf{Top}) and tool evolution (\textbf{Bottom}) when tools are partially missing on Earth-Bench and OpenEarthAgent-Bench, respectively.
    }
    \label{fig:evolution_trend}
\end{figure}

\begin{figure}[t]
    \centering
    \includegraphics[width=\columnwidth]{
        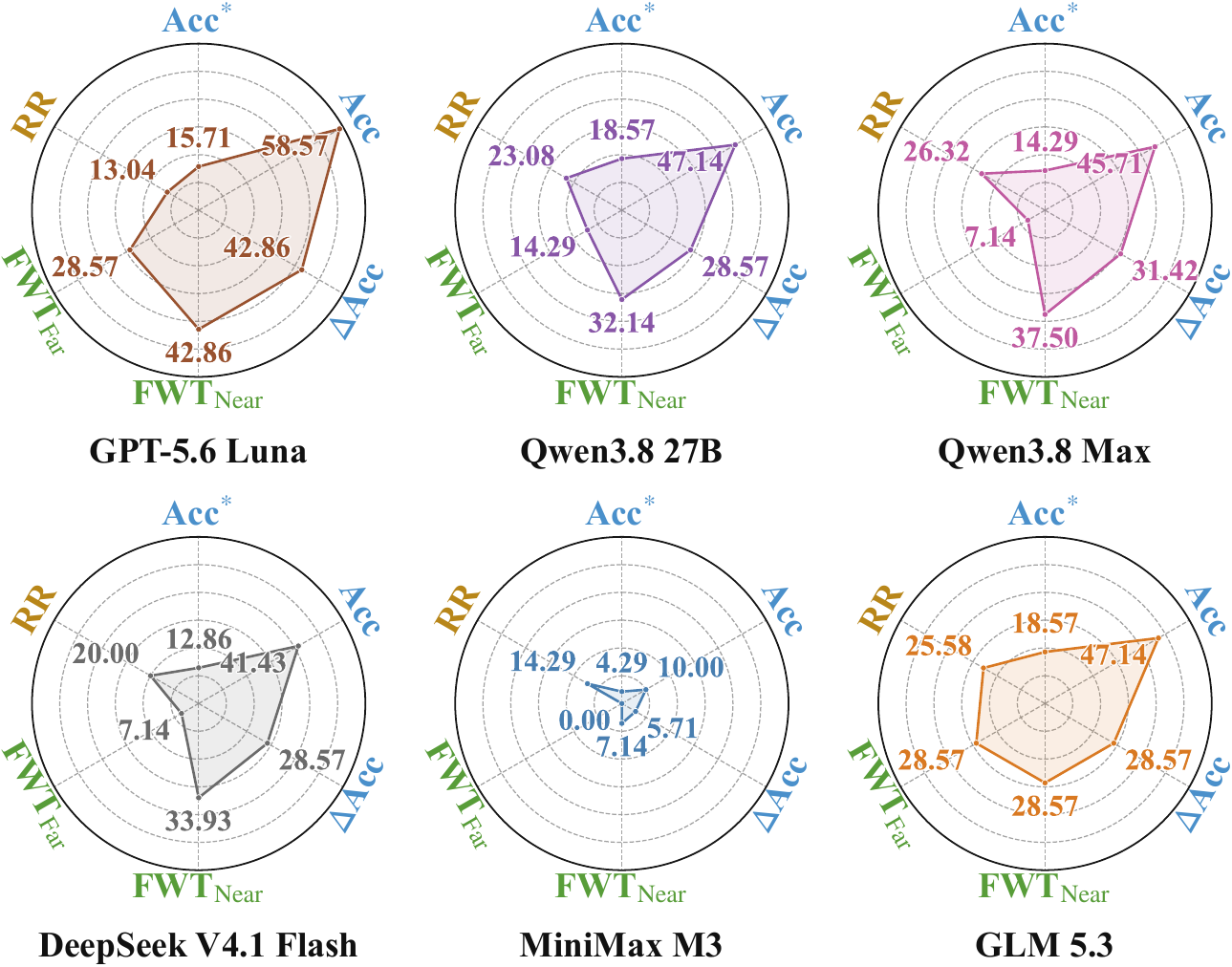
    }
    \caption{
        Generality across foundation models on Earth-Bench under missing tools. All runs use full RemoteEvolver with identical settings except for the base model.
    }
    \label{fig:model-radar}
\end{figure}

\subsection{Evolution Trends}
\label{sec:evolution_trend}
As shown in Fig.~\ref{fig:evolution_trend}, while the overall accuracy initially rises, subsequent checkpoints reveal a non-monotonic trend. Specifically, while near-domain transfer $\mathrm{FWT}_{\mathrm{Near}}$ remains largely positive, far-domain transfer $\mathrm{FWT}_{\mathrm{Far}}$ suffers from severe fluctuations, sometimes dropping below 0 on OpenEarthAgent-Bench. Simultaneously, the retention rate abnormally increases, indicating the agent forgets previously mastered tasks. These observations highlight a critical insight: \textbf{simply expanding the harness does not guarantee stable performance growth}. Thus, we conduct a deep analysis and attribute this non-monotonic degradation to a cascading effect across three levels of the agent's expanded action space: tool exploration, skill retrieval, and workflow execution, as illustrated in Appendix.

First, over-exploration at the tool level. While recovering missing tools equips the agent with essential capabilities, it introduces increased decision noise into planning. Consequently, the agent tends to over-complicate originally simple tasks by repeatedly invoking irrelevant newly-acquired tools, ultimately leading to failure via exhausted execution rounds or the gradual accumulation of intermediate errors.

Second, negative transfer at the skill level. Even when the correct tools are selected, distilled skills encapsulating general rules may possess vague applicability boundaries. This ambiguity causes the system to retrieve seemingly relevant historical skills that actually conflict with the constraints of the new task, thereby misguiding the workflow.

Finally, execution loss at the planning level. Even if an appropriate skill is retrieved, the agent may still fail to strictly follow its guidelines. During reasoning processes, the internal planner frequently overlooks critical constraints within the skill document, resulting in a disconnect between semantic comprehension and actual execution.

% To intuitively illustrate these cascading challenges, we provide concrete execution trajectories in Appendix~\ref{sec:attribution}.

\subsection{Further Analysis}
\label{sec:further_analysis}
To provide deeper insights into the mechanisms and practical advantages of RemoteEvolver, we further systematically explore the following core questions.

\noindent
\textbf{Does external harness evolution generalize universally across diverse foundation models?}
As illustrated in Fig.~\ref{fig:model-radar}, we extend our framework to 6 diverse foundation models under the \textit{Tool-Missing} setting. Regardless of their varying parameter sizes and initial capabilities, all evaluated models consistently expand their capabilities, achieving positive accuracy gains and improved near-domain transfer. 

Notably, a model's initial zero-shot performance does not strictly dictate its evolutionary potential. For instance, GPT-5.6 Luna does not possess the highest initial accuracy, trailing the 18.57$\%$ of GLM 5.3. However, it exhibits exceptional capability in assimilating external tools and experiences, securing a massive 42.86$\%$ boost to reach the highest absolute accuracy of 58.57$\%$. These results confirm that external harness evolution is a universally applicable paradigm that thoroughly decouples continuous system growth from the inherent limits of static internal weights.

\begin{figure}[h]
    \centering
    \includegraphics[width=\columnwidth]{
        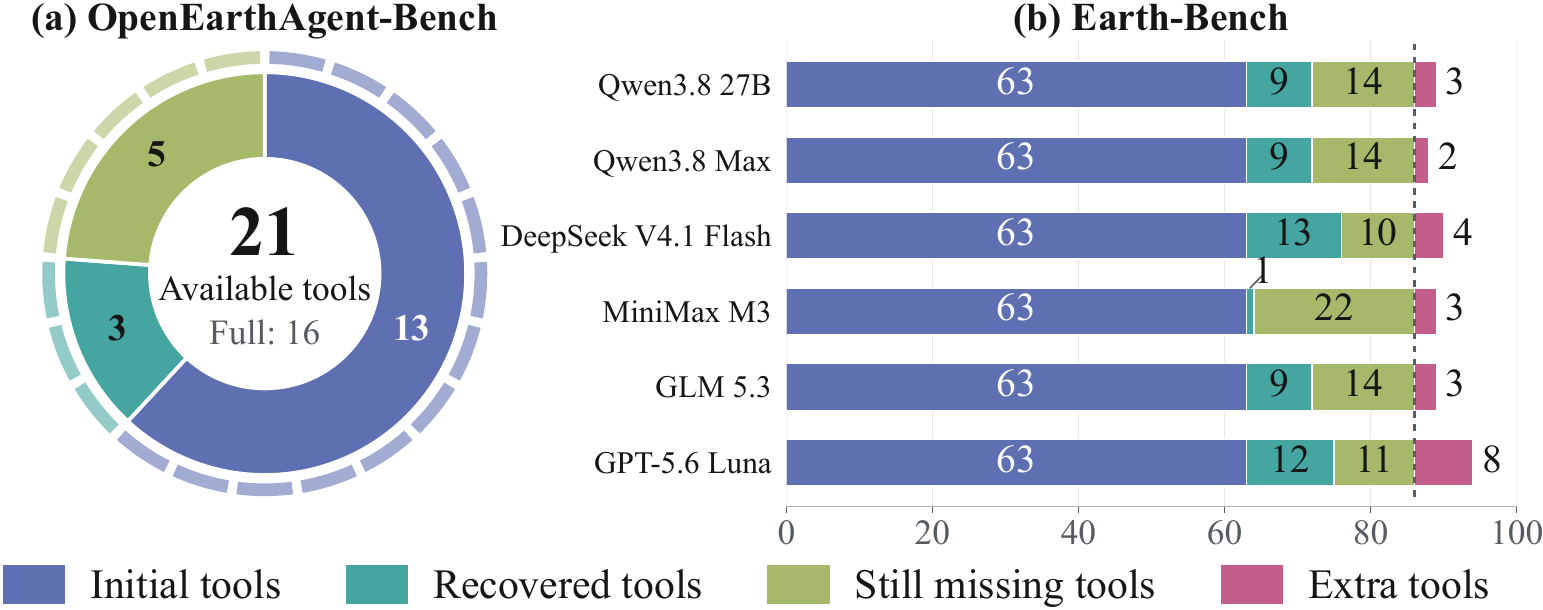
    }
    \caption{
        Tool statistics under the \textit{Tool-Missing} setting. The final toolset composition is categorized into initial tools, successfully recovered tools, tools that remain missing, and newly acquired extra tools not present in the original full set.
    }
    \label{fig:tool_statistics}
\end{figure}

\noindent
\textbf{How do agents dynamically expand and recover toolsets under missing conditions?}
We further analyze the final toolset composition after evolution. As shown in Fig.~\ref{fig:tool_statistics}, the agents exhibit robust dynamic expansion capabilities, transitioning from passive recovery to proactive exploration.

Tool recovery capacity is highly correlated with the model's final task performance. On Earth-Bench, where all models are restricted to 63 initial tools, powerful foundation models, such as DeepSeek V4.1 Flash and GPT-5.6 Luna, successfully recovered 13 and 12 core tools. Conversely, models like MiniMax M3 only managed to recover 1 tool, leaving 22 tools persistently missing, which directly explains its suboptimal accuracy observed in Q1.

The system also exhibits open-ended discovery beyond predefined boundaries. The agents did not confine their exploration to the predefined full set of tools. As illustrated in Fig.~\ref{fig:tool_statistics}, multiple models successfully acquired extra tools absent from the original full set. Most notably, the top-performing GPT-5.6 Luna autonomously introduced 8 extra tools to address the unique demands of complex tasks. This phenomenon provides evidence that RemoteEvolver endows agents with open-ended toolset expansion capabilities, thoroughly liberating them from the constraints of static tool libraries to achieve true on-demand growth.

\begin{figure}[t]
    \centering
    \includegraphics[width=\columnwidth]{
        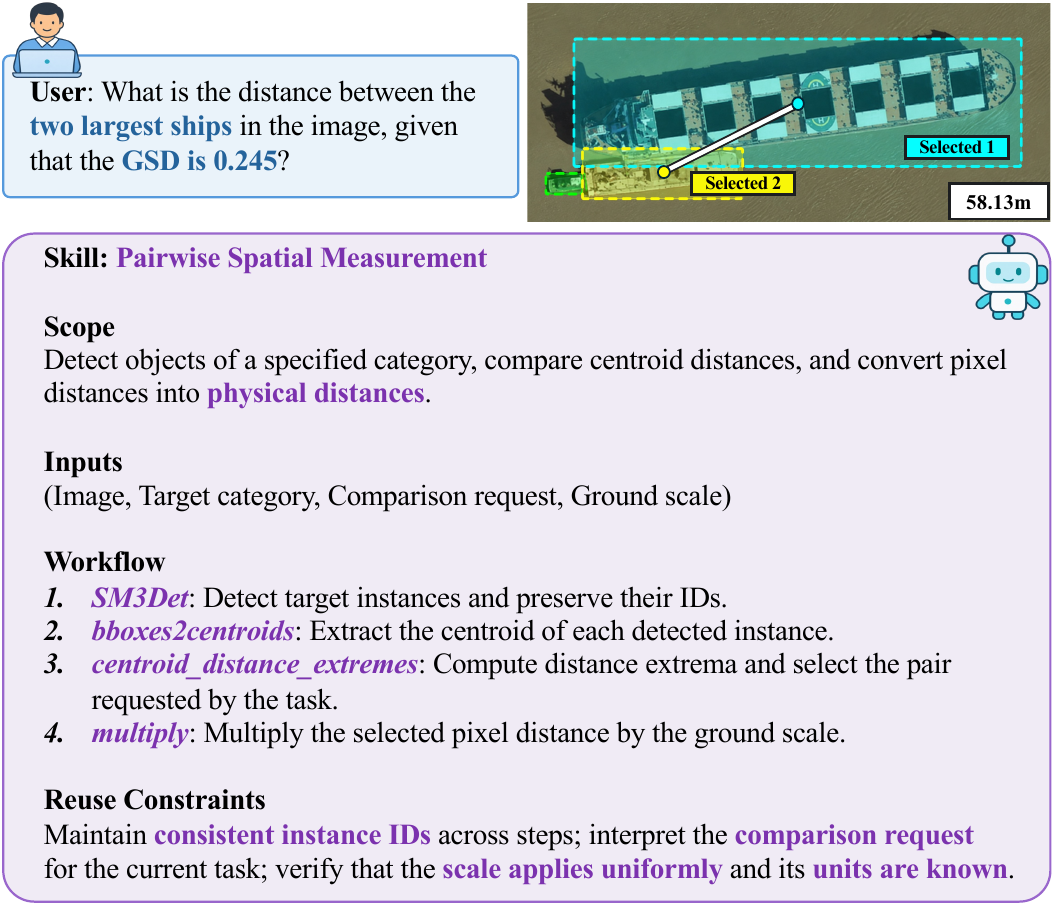
    }
    \caption{
        Example of a spatial measurement task and its associated distilled skill document.
    }
    \label{fig:skill}
\end{figure}

\begin{figure}[b]
    \centering
    \includegraphics[width=\columnwidth]{
        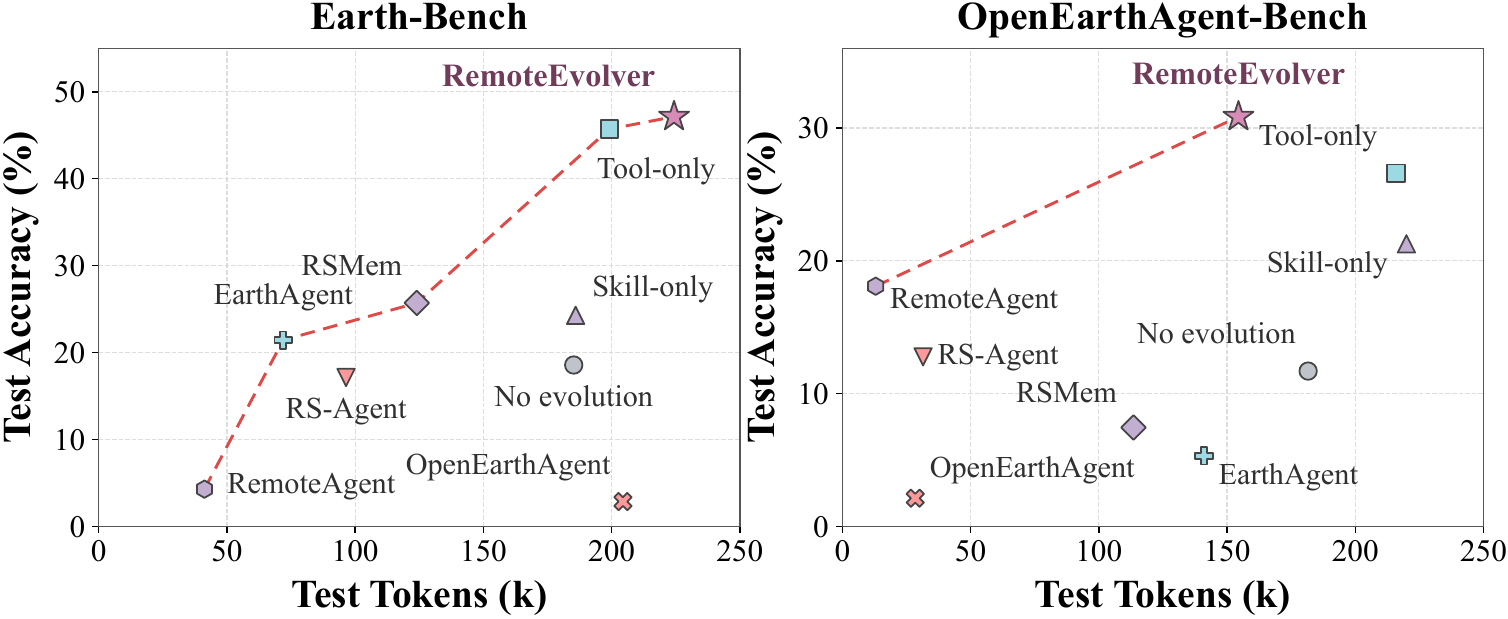
    }
    \caption{
        System efficiency and Pareto front analysis of remote sensing agents. The scatter plots compare the test accuracy against the token consumption (in thousands) during the testing phase.
    }
    \label{fig:pareto}
\end{figure}

\begin{figure*}[t]
    \centering
    \includegraphics[width=\textwidth]{
    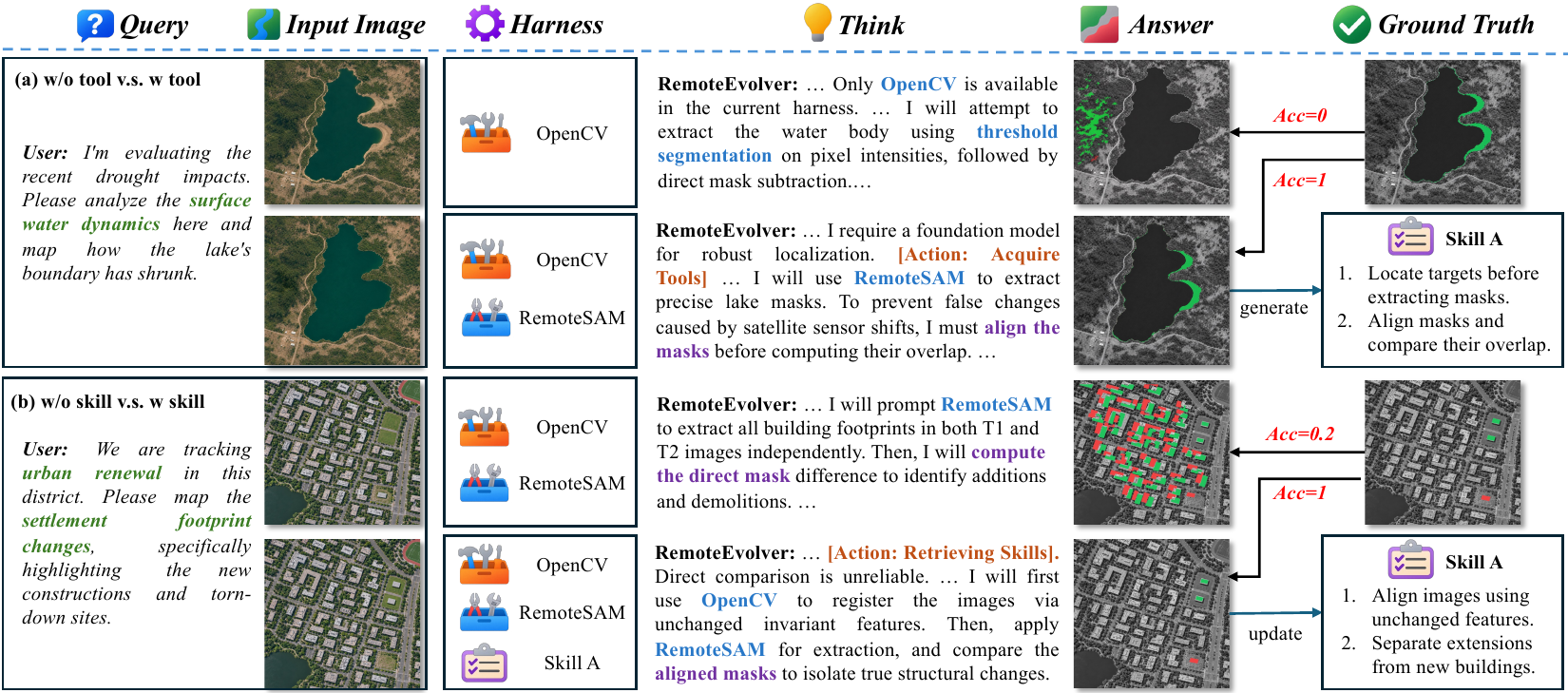
    }
    \caption{
        Examples of RemoteEvolver's execution trajectories. (a) Without vs. with tool evolution for mapping lake boundary shrinkage. The agent autonomously acquires RemoteSAM to replace inadequate OpenCV threshold segmentation. (b) Without vs. with skill evolution for tracking building contour changes. The agent retrieves historical skills to align images before comparing masks.
    }
    \label{fig:case_study}
\end{figure*}

\noindent
\textbf{Is the computational overhead of harness evolution justified by Pareto efficiency?}
Continuously invoking external tools and retrieving historical skills inherently increases prompt lengths and execution rounds. To verify whether this computational overhead is acceptable, we map test accuracy against token consumption in Fig.~\ref{fig:pareto}. Although the full RemoteEvolver naturally consumes more tokens than static baselines like EarthAgent or RSMem, it establishes a distinctly dominant Pareto front across both benchmarks.

\noindent
\textbf{How are successful execution workflows distilled into structured reusable skills?}
Unlike traditional methods that attempt to internalize experience into model weights, RemoteEvolver distills verified successful execution trajectories into human-readable and structured skill documents. As illustrated in Fig.~\ref{fig:skill}, a successfully verified spatial measurement workflow is preserved with explicitly defined scopes, input formats, and critical reuse constraints. This structural clarity guarantees reliable retrieval and precise execution in subsequent tasks.

\noindent
\textbf{How does RemoteEvolver overcome capability gaps and avoid error propagation in complex scenarios?}
As illustrated in Fig.~\ref{fig:case_study}, we demonstrate how RemoteEvolver tackles complex remote sensing tasks through concrete execution trajectories.
In Fig.~\ref{fig:case_study}(a), when tasked with evaluating drought impacts, the initial agent relies solely on basic OpenCV thresholding and fails due to inaccurate lake boundary extraction. By enabling tool evolution, the agent autonomously acquires \texttt{RemoteSAM} to precisely map the water boundary and successfully completes the task.

In Fig.~\ref{fig:case_study}(b), when tracking urban building changes without historical experience, the agent directly computes mask differences, leading to massive false detections caused by sensor shifts. By applying historical skills, the agent learns to align images using invariant features before extracting masks, isolating structural changes and achieving accurate task completion. These cases demonstrate that tool expansion breaks through perceptual ceilings, while skill consolidation regularizes complex reasoning paths, jointly ensuring robust agent execution in complex environments.

%% file: 5_limitation.tex
\section{Limitation}
While RemoteEvolver demonstrates promising progress, there remain controllable limitations to address. As revealed in the evolution trend analysis in Section~\ref{sec:further_analysis}, unconstrained harness evolution can sometimes lead to non-monotonic performance fluctuations, such as degraded far-domain transfer and the forgetting of previously mastered tasks. This is primarily driven by a cascading effect caused by over-exploration at the tool level, negative transfer at the skill level, and execution loss at the planning level. Future work will focus on incorporating robust skill verification mechanisms and efficient evolution pruning strategies. These improvements aim to better control computational overhead and suppress decision noise while ensuring continuous capability growth, ultimately achieving more stable, reliable, and monotonic system evolution over time.

%% file: 6_Conclusion.tex
\section{Conclusion}
In this paper, we explore the continuous self-improvement of remote sensing agentic systems when encountering capability gaps. To advance this direction, we propose RemoteEvolver, a novel harness-centric self-evolving paradigm. This framework provides the community with a practical training-free solution that enables agents to seamlessly integrate emerging external specialized models and reuse accumulated execution experience across tasks. Our evaluations show that RemoteEvolver establishes a new state-of-the-art on Earth-Bench and OpenEarthAgent-Bench under both complete and constrained tool settings. Ultimately, we hope this work catalyzes a paradigm shift in remote sensing agent community, moving beyond reliance on static, pre-integrated toolsets toward developing intelligent systems equipped with continuous self-growing capabilities.

%% file: 7_LLM.tex
\section{LLM Usage Statement}
ChatGPT, Codex, and Claude Code were used exclusively for manuscript language refinement, author-reviewed code generation, and experimental progress monitoring. They were not used to generate research ideas, figures, citations, data, or conclusions. The human authors retained full control and ultimate responsibility for all scientific judgments, manuscript content, and experimental results.

%% file: main.bib
@String(CVPR= {IEEE Conf. Comput. Vis. Pattern Recog.})

@String(AAAI = {AAAI})

@String(CVPR  = {CVPR})

@inproceedings{yao2026remotereasoner,
  title={Remotereasoner: Towards unifying geospatial reasoning workflow},
  author={Yao, Liang and Liu, Fan and Lu, Hongbo and Zhang, Chuanyi and Min, Rui and Xu, Shengxiang and Di, Shimin and Peng, Pai},
  booktitle={Proceedings of the AAAI Conference on Artificial Intelligence},
  volume={40},
  number={14},
  pages={11883--11891},
  year={2026}
}

@article{tang2026intelligent,
  title={Intelligent remote sensing agents: A survey},
  author={Tang, Jiaqi and Yan, Yingying and Wang, Qianzhou and Xia, Yuyang and Geng, Botong and Chen, Jianmin and Ma, Ke and Zhai, Youyang and He, Qingfeng and Shao, Weigeng and others},
  year={2026},
  publisher={Technical Report, 2026c. URL https://github. com/PolyX-Research/Awesome~…}
}

@article{zhao2026openearth,
  title={Openearth-agent: From tool calling to tool creation for open-environment earth observation},
  author={Zhao, Sijie and Liu, Feng and Zhang, Xueliang and Chen, Hao and Gu, Xinyu and Jiang, Zhe and Ling, Fenghua and Fei, Ben and Zhang, Wenlong and Wang, Junjue and others},
  journal={arXiv preprint arXiv:2603.22148},
  year={2026}
}

@article{dai2026experience,
  title={Experience-driven multi-agent systems are training-free context-aware earth observers},
  author={Dai, Pengyu and Xuan, Weihao and Wang, Junjue and Chen, Hongruixuan and Song, Jian and Ou, Yafei and Yokoya, Naoto},
  journal={arXiv preprint arXiv:2602.02559},
  year={2026}
}

@article{talemi2026agentic,
  title={Agentic ai in remote sensing: Foundations, taxonomy, and emerging systems},
  author={Talemi, Niloufar Alipour and Boone, Julia and Afghah, Fatemeh},
  journal={arXiv preprint arXiv:2601.01891},
  year={2026}
}

@inproceedings{wu2026rsmem,
  title={RSMeM: Knowledge-Enhanced Memory Evolution for Remote Sensing Agents with Systematic Evaluation},
  author={Wu, Bingxian and Zhang, Yu and Guo, Zonghao and Liu, Tang and Qian, Chen and Lu, Yuxiang and Du, Xingbo and Li, Yanghao and Zhang, Yidan and Chen, Chi and others},
  booktitle={Proceedings of the 64th Annual Meeting of the Association for Computational Linguistics (Volume 1: Long Papers)},
  pages={32899--32915},
  year={2026}
}

@article{chen2025cangling,
  title={Cangling-knowflow: A unified knowledge-and-flow-fused agent for comprehensive remote sensing applications},
  author={Chen, Zhengchao and Wang, Haoran and Yao, Jing and Zhang, Jianshe and Ghamisi, Pedram and Zhou, Jun and Atkinson, Peter M and Zhang, Bing},
  journal={arXiv preprint arXiv:2512.15231},
  year={2025}
}

@article{xiao2026geoforge,
  title={GeoForge: Non-Parametric Self-Evolving Agents for Earth-Observation Reasoning},
  author={Xiao, Xin and Zhong, Jiang and Zhu, Junnan and Feng, Yingchao and Wang, Peijin and Zhang, Yidan and Wei, Kaiwen},
  journal={arXiv preprint arXiv:2608.10494},
  year={2026}
}

@article{yao2026remoteagent,
  title={Remoteagent: Bridging vague human intents and earth observation with rl-based agentic mllms},
  author={Yao, Liang and Xu, Shengxiang and Liu, Fan and Zhang, Chuanyi and Yao, Bishun and Min, Rui and Li, Yongjun and Ouyang, Chaoqian and Di, Shimin and Zhang, Min-Ling},
  journal={arXiv preprint arXiv:2604.07765},
  year={2026}
}

@inproceedings{feng2026earth,
  title={Earth-agent: Unlocking the full landscape of earth observation with agents},
  author={Feng, Peilin and Lv, Zhutao and Ye, Junyan and Wang, Xiaolei and Huo, Xinjie and Yu, Jinhua and Xu, Wanghan and Zhang, Wenlong and Bai, Lei and He, Conghui and others},
  booktitle={International Conference on Learning Representations},
  volume={2026},
  pages={55700--55764},
  year={2026}
}

@article{xu2026rs,
  title={Rs-agent: Automating remote sensing tasks through intelligent agent},
  author={Xu, Wenjia and Yu, Zijian and Mu, Boyang and Wang, Jiuniu and Wei, Zhiwei and Peng, Mugen},
  journal={Science China Information Sciences},
  volume={69},
  number={8},
  pages={180302},
  year={2026},
  publisher={Springer}
}

@inproceedings{shabbir2026openearthagent,
  title={Openearthagent: A unified framework for tool-augmented geospatial agents},
  author={Shabbir, Akashah and Umer Sheikh, Muhammad and Akhtar Munir, Muhammad and Debary, Hiyam and Fiaz, Mustansar and Zaigham Zaheer, Muhammad and Fraccaro, Paolo and Shahbaz Khan, Fahad and Haris Khan, Muhammad and Xiang Zhu, Xiao and others},
  booktitle={European Conference on Computer Vision},
  pages={172--191},
  year={2026},
  organization={Springer}
}

@misc{qwen38,
    title = {Qwen3.8-Max: A New Bar for Coding and Cowork},
    url = {https://qwen.ai/blog?id=qwen3.8},
    author = {{Qwen Team}},
    month = {August},
    year = {2026}
}

@article{xu2026deepseek,
  title={Deepseek-v4. 1-flash: Pushing the limits of kv cache compression},
  author={Xu, Anyi and Li, B and Lin, Bangcai and Xue, Bing and Xian, BingCheng and Xu, Bingzheng and Wu, Bochao and Zhang, Bowei and Deng, Boyi and Yu, CC and others},
  journal={arXiv preprint arXiv:2609.19969},
  year={2026}
}

@article{lai2026minimax,
  title={Minimax sparse attention},
  author={Lai, Xunhao and Xu, Weiqi and Yang, Yufeng and Chen, Qiaorui and Xu, Yang and Zeng, Lunbin and Li, Xiaolong and Sun, Haohai and Zhu, Haichao and Zhang, Vito and others},
  journal={arXiv preprint arXiv:2606.13392},
  year={2026}
}

@article{zeng2026glm,
  title={Glm-5: from vibe coding to agentic engineering},
  author={Zeng, Aohan and Lv, Xin and Hou, Zhenyu and Du, Zhengxiao and Zheng, Qinkai and Chen, Bin and Yin, Da and Ge, Chendi and Huang, Chenghua and Xie, Chengxing and others},
  journal={arXiv preprint arXiv:2602.15763},
  year={2026}
}

@article{lv2026earth,
  title={Earth-Agent-Pro: Towards Real-World Full-Chain Earth Observation with Agents},
  author={Lv, Zhutao and Dang, Chenhao and Feng, Yi and Gong, Yanpei and Wang, Xiaolei and Ye, Junyan and He, Conghui and Li, Weijia},
  journal={arXiv preprint arXiv:2609.12533},
  year={2026}
}

@article{cong2022satmae,
  title={Satmae: Pre-training transformers for temporal and multi-spectral satellite imagery},
  author={Cong, Yezhen and Khanna, Samar and Meng, Chenlin and Liu, Patrick and Rozi, Erik and He, Yutong and Burke, Marshall and Lobell, David and Ermon, Stefano},
  journal={Advances in Neural Information Processing Systems},
  volume={35},
  pages={197--211},
  year={2022}
}

@article{fuller2023croma,
  title={Croma: Remote sensing representations with contrastive radar-optical masked autoencoders},
  author={Fuller, Anthony and Millard, Koreen and Green, James},
  journal={Advances in Neural Information Processing Systems},
  volume={36},
  pages={5506--5538},
  year={2023}
}

@inproceedings{guo2024skysense,
  title={Skysense: A multi-modal remote sensing foundation model towards universal interpretation for earth observation imagery},
  author={Guo, Xin and Lao, Jiangwei and Dang, Bo and Zhang, Yingying and Yu, Lei and Ru, Lixiang and Zhong, Liheng and Huang, Ziyuan and Wu, Kang and Hu, Dingxiang and others},
  booktitle={2024 IEEE/CVF Conference on Computer Vision and Pattern Recognition (CVPR)},
  pages={27662--27673},
  year={2024},
  organization={IEEE}
}

@inproceedings{astruc2025anysat,
  title={Anysat: One earth observation model for many resolutions, scales, and modalities},
  author={Astruc, Guillaume and Gonthier, Nicolas and Mallet, Clement and Landrieu, Loic},
  booktitle={2025 IEEE/CVF Conference on Computer Vision and Pattern Recognition (CVPR)},
  pages={19530--19540},
  year={2025},
  organization={IEEE}
}

@article{liu2024remoteclip,
  title={Remoteclip: A vision language foundation model for remote sensing},
  author={Liu, Fan and Chen, Delong and Guan, Zhangqingyun and Zhou, Xiaocong and Zhu, Jiale and Ye, Qiaolin and Fu, Liyong and Zhou, Jun},
  journal={IEEE Transactions on Geoscience and Remote Sensing},
  volume={62},
  pages={1--16},
  year={2024},
  publisher={IEEE}
}

@inproceedings{kuckreja2024geochat,
  title={Geochat: Grounded large vision-language model for remote sensing},
  author={Kuckreja, Kartik and Danish, Muhammad Sohail and Naseer, Muzammal and Das, Abhijit and Khan, Salman and Khan, Fahad Shahbaz},
  booktitle={2024 IEEE/CVF Conference on Computer Vision and Pattern Recognition (CVPR)},
  pages={27831--27840},
  year={2024},
  organization={IEEE}
}

@inproceedings{soni2025earthdial,
  title={Earthdial: Turning multi-sensory earth observations to interactive dialogues},
  author={Soni, Sagar and Dudhane, Akshay and Debary, Hiyam and Fiaz, Mustansar and Munir, Muhammad Akhtar and Danish, Muhammad Sohail and Fraccaro, Paolo and Watson, Campbell D and Klein, Levente J and Khan, Fahad Shahbaz and others},
  booktitle={2025 IEEE/CVF Conference on Computer Vision and Pattern Recognition (CVPR)},
  pages={14303--14313},
  year={2025},
  organization={IEEE}
}

@inproceedings{yao2025remotesam,
  title={Remotesam: Towards segment anything for earth observation},
  author={Yao, Liang and Liu, Fan and Chen, Delong and Zhang, Chuanyi and Wang, Yijun and Chen, Ziyun and Xu, Wei and Di, Shimin and Zheng, Yuhui},
  booktitle={Proceedings of the 33rd ACM international conference on multimedia},
  pages={3027--3036},
  year={2025}
}

@article{zhou2024geoground,
  title={Geoground: A unified large vision-language model for remote sensing visual grounding},
  author={Zhou, Yue and Lan, Mengcheng and Li, Xiang and Feng, Litong and Ke, Yiping and Jiang, Xue and Li, Qingyun and Yang, Xue and Zhang, Wayne},
  journal={arXiv preprint arXiv:2411.11904},
  year={2024}
}

@article{shabbir2025geopixel,
  title={Geopixel: Pixel grounding large multimodal model in remote sensing},
  author={Shabbir, Akashah and Zumri, Mohammed and Bennamoun, Mohammed and Khan, Fahad S and Khan, Salman},
  journal={arXiv preprint arXiv:2501.13925},
  year={2025}
}

@article{xu2025robustflow,
  title={Robustflow: Towards robust agentic workflow generation},
  author={Xu, Shengxiang and Zhang, Jiayi and Di, Shimin and Luo, Yuyu and Yao, Liang and Liu, Hanmo and Zhu, Jia and Zhang, Min-Ling},
  journal={arXiv preprint arXiv:2509.21834},
  year={2025}
}

@article{hu2026ringmo,
  title={RingMo-Agent: A unified remote sensing foundation model for multi-platform and multi-modal reasoning},
  author={Hu, Huiyang and Wang, Peijin and Feng, Yingchao and Wei, Kaiwen and Yin, Wenxin and Diao, Wenhui and Wang, Mengyu and Bi, Hanbo and Kang, Kaiyue and Ling, Tong and others},
  journal={IEEE Transactions on Pattern Analysis and Machine Intelligence},
  year={2026},
  publisher={IEEE}
}

@inproceedings{kao2026towards,
  title={Towards llm agents for earth observation},
  author={Kao, Chia Hsiang and Zhao, Wenting and Lam, Cheryl and Umap, Aarush and Revankar, Shreelekha and Speas, Samuel and Bhagat, Snehal and Datta, Rajeev and Phoo, Cheng Perng and Mall, Utkarsh and others},
  booktitle={Findings of the Association for Computational Linguistics: ACL 2026},
  pages={2597--2611},
  year={2026}
}

@inproceedings{weiss2025mapbot,
  title={Mapbot: A multi-modal agent for geospatial analysis},
  author={Weiss, Martin and Rahaman, Nasim and Pal, Chris},
  booktitle={Proceedings of the 24th International Conference on Autonomous Agents and Multiagent Systems},
  pages={3059--3061},
  year={2025}
}

@article{yao2026remotezero,
  title={RemoteZero: Geospatial Reasoning with Zero Human Annotations},
  author={Yao, Liang and Liu, Fan and Xu, Shengxiang and Zhang, Chuanyi and Min, Rui and Di, Shimin and Zheng, Yuhui},
  journal={arXiv preprint arXiv:2605.04451},
  year={2026}
}

@inproceedings{xin2026segearth,
  title={Segearth-r2: Towards comprehensive language-guided segmentation for remote sensing images},
  author={Xin, Zepeng and Li, Kaiyu and Chen, Luodi and Li, Wanchen and Yuchen, Xiao and Qiao, Hui and Zhang, Weizhan and Meng, Deyu and Cao, Xiangyong},
  booktitle={Proceedings of the IEEE/CVF Conference on Computer Vision and Pattern Recognition},
  pages={13199--13210},
  year={2026}
}

@article{guo2026drevo,
  title={DREvo: Distilling Recalibrated Historical Experience for Harness Self-Evolution},
  author={Guo, Hanghui and Shi, Weijie and Chen, Zhangze and Xu, Shengxiang and Wang, Yishu and Zhang, Yimei and Ni, Wangze and Zhu, Jia and Di, Shimin},
  journal={arXiv preprint arXiv:2607.26722},
  year={2026}
}

@article{ren2026flowevo,
  title={Flowevo: Self-evolving agents through the co-evolution of workflows and executable skills},
  author={Ren, Zeyu and Yue, Ling and Li, Ran and Wang, Yishu and Xu, Shengxiang and Liu, Hanmo and Pan, Shaowu and Di, Shimin},
  year={2026}
}

@article{mu2026hirs,
  title={HiRS-Agent: A Hierarchical Multi-Agent System for Reliable Long-Horizon Remote Sensing Task Solving},
  author={Mu, Boyang and Wei, Zhiwei and Peng, Mugen and Xu, Wenjia},
  journal={arXiv preprint arXiv:2608.30672},
  year={2026}
}
